\documentclass{article}

\usepackage{PRIMEarxiv}

\usepackage[utf8]{inputenc}
\usepackage[T1]{fontenc}   
\usepackage{hyperref}      
\usepackage{url}           
\usepackage{booktabs}      
\usepackage{amsfonts}      
\usepackage{nicefrac}      
\usepackage{microtype}     
\usepackage{lipsum}
\usepackage{fancyhdr}      
\usepackage{graphicx}      
\graphicspath{{media/}}    

\usepackage{longtable}
\usepackage{amssymb}
\usepackage{amsmath}
\usepackage{bm}
\usepackage{subfigure}

\usepackage{natbib}
\setcitestyle{authoryear, open={(}, close={)}}

\title{Predicting Corporate Risk by Jointly Modeling Company Networks and Dialogues in Earnings Conference Calls}

\author{
  Yunxin Sang \\
  Shanghai Jiao Tong University \\
  \texttt{sangyunxin@gmail.com} \\
   \And
  Yang Bao \\
  Shanghai Jiao Tong University \\
  \texttt{baoyang@sjtu.edu.cn} \\
}

\begin{document}
\maketitle

\begin{abstract}
Earnings conference calls are significant information events for volatility forecasting, which is essential for financial risk management and asset pricing. Although some recent volatility forecasting models have utilized the textual content of conference calls, the dialogue structures of conference calls and company relationships are almost ignored in extant literature. To bridge this gap, we propose a new model called Temporal Virtual Graph Neural Network (TVGNN) for volatility forecasting by jointly modeling conference call dialogues and company networks. Our model differs from existing models in several important ways. First, we propose to exploit more dialogue structures by encoding position, utterance, speaker role, and Q\&A segments. Second, we propose to encode the market states for volatility forecasting by extending the Gated Recurrent Units (GRU). Third, we propose a new method for constructing temporal company networks in which the messages can only flow from temporally preceding to successive nodes, and extend the Graph Attention Networks (GAT) for modeling company relationships. We collect conference call transcripts of S\&P500 companies from 2008 to 2019, and construct a dataset of conference call dialogues with additional information on dialogue structures and company networks. Empirical results on our dataset demonstrate the superiority of our model over competitive baselines for volatility forecasting. We also conduct supplementary analyses to examine the effectiveness of our model's key components and interpretability.
\end{abstract}


\section{Introduction} \label{sec:introduction}

Volatility is a statistical measure of variation in a stock's returns over time, which plays an essential role in financial risk management and asset pricing \citep{poon2003forecasting}. Due to its importance for financial risk assessment, volatility forecasting has attracted attention from various stakeholders such as researchers, investors, analysts, and other market participants. Volatility forecasting models are traditionally built using the related numerical feature variables (e.g., historical volatility), but recent studies have shown that the textual information of corporate disclosures could provide incremental information over the numerical variables for volatility forecasting \citep{kogan2009predicting}. In this line of research, annual reports and earnings conference calls are the two most examined types of corporate disclosures for financial risk prediction \citep{matsumoto2011makes,bao2014simultaneously}. Compared with the mandated annual reports, earnings conference calls have become an increasingly important form of voluntary disclosure that is more casual and spontaneous. Specifically, public companies usually hold quarterly conference calls to communicate their financial and operational results to interested parties such as investors and analysts. Each conference call typically contains two segments, i.e., the presentation and Q\&A (Question and Answer). The management executives (e.g., CEO, CFO, or other executives) will first provide their interpretation of the firm's performance in the presentation session, and the buy-side or sell-side analysts could ask questions, request more details, and perhaps question management's interpretation in a follow-up Q\&A session. Due to the less constrained fashion and direct interaction between managers and analysts, earnings conference calls have been recognized as significant information events to the market \citep{matsumoto2011makes}. For example, the disclosure of a higher-than-expected loss by Advanced Micro Devices, Inc (AMD). during its earnings conference call in the first quarter of 2017 caused its shares to plunge 16.1\% after the conference call.

Although the extant literature has documented the information value of earnings conference calls and attempted to utilize their textual contents for volatility forecasting, two essential types of information, i.e., the dialogue structures and company relationships, are almost ignored in extant literature. On the one hand, the earnings conference call could affect the risk perceptions of market participants not only by what is said (i.e., utterance) but also by who said it (e.g., speaker role) and how it is said (e.g., position and segment information) \citep{qin2019what,theil2019profet}. For example, the management executives who hold inside information might be reluctant to disclose negative information during the presentation segment or dodge questions during the Q\&A segment, while the analysts outside the company might ask acute questions based on their own or the institution's position, which often reflects the market's attitude towards the company. On the other hand, the relationships between companies also play a crucial role in volatility forecasting because the risk could transmit from one company to another in the company network \citep{sawhney2020voltage}. For example, the bankruptcy of Lehman Brothers (the fourth-largest U.S. investment bank at the time) quickly triggered a chain reaction leading to a subsequent global financial crisis in 2008.

To address the aforementioned issues, we propose a new model called Temporal Virtual Graph Neural Network (TVGNN) for volatility forecasting by jointly modeling conference call dialogues and company networks. Our model differs from existing models in several important ways:
\begin{enumerate}
\item We propose to exploit more dialogue structures by encoding position, utterance, speaker role, and Q\&A segments.
\item We propose to encode the market states for volatility forecasting by extending the Gated Recurrent Units (GRU).
\item We propose a new method for constructing temporal company networks in which the messages can only flow from temporally preceding to successive nodes, and extend the Graph Attention Networks (GAT) for modeling company relationships.
\end{enumerate}

We collect conference call transcripts of S\&P500 companies from 2008 to 2019, and construct a dataset of conference call dialogues with additional information on dialogue structures and company networks. Empirical results on our dataset demonstrate the superiority of our model over competitive baselines for volatility forecasting. We also conduct supplementary analyses to examine the effectiveness of our model's key components and interpretability.

\section{Related Work}\label{sec:literature}
This study is related to two main areas, including graph neural networks and financial risk prediction.

\subsection{Graph Neural Network}
Graph data is ubiquitous and useful for many real-world applications. But the non-Euclidean graph data contains rich relational information and cannot be well handled by traditional neural networks. To tackle this problem, researchers proposed graph neural networks (GNN) to learn better representations of graphs via message passing. The early GNN models \citep{gori2005new,scarselli2008graph} update the node embedding iteratively based on Banach's fixed point theorem, which is inefficient. \citet{bruna2013spectral} proposed the graph convolutional networks (GCN), and \citet{kipf2016semi} improved the GCN's computational efficiency by 
computing convolutional kernels with Chebyshev polynomials and renormalization techniques. Recently, a variety of graph neural networks are proposed, such as Graph Isomorphism Network \citep{xu2018powerful}, Graph Attention Network (GAT) \citep{velivckovic2017graph}, and Graphormer \citep{ying2021transformers}. These standard GNN models focused on static graphs with fixed nodes and edges, whereas many practical applications involve dynamic graphs with changing nodes or edges. A simple way to deal with dynamic graphs is to convert them into static graphs. For example, \citet{liben2007link} converted dynamic graphs to static graphs by adding the adjacency matrices of graphs at different time steps, and \citet{hisano2018semi} modeled the temporal information using the formation and dissolution matrices of previous $k$ time steps. Aside from that, it is more natural to separately model the graphs at different time steps and then aggregate the results. For example, \citet{yao2016link} used graph neural networks to model the graph snapshots at each time step, and then weighted these snapshots based on their time difference from the current time step. Departing from existing models, we propose a new graph construction method and extend the GAT model for handling temporal company network.

\subsection{Finanical Risk Prediction}
Volatility forecasting is an essential task for risk assessment in financial markets. Traditional volatility forecasting models are built solely based on the numerical feature variables, but recent studies have shown that the textual information of corporate disclosures could provide incremental information over the numerical variables \citep{kogan2009predicting}. In this line of research, annual reports and earnings conference calls are the two most examined types of corporate disclosures for financial risk prediction \citep{matsumoto2011makes,bao2014simultaneously}. For example, \citet{kogan2009predicting} are among the first to predict stock return volatility using corporate annual reports. More recent studies have demonstrated that the earnings conference calls are more informative due to their less constrained fashion and direct interactions between managers and analysts. For example, \citet{qin2019what} proposed a multi-modal deep regression model to capture textual and audio features in earnings conference calls for financial risk prediction. \citet{theil2019profet} utilized a hierarchical recurrent neural network to model the textual features of the presentation and Q\&A segments separately. \citet{li2020maec} collected textual and audio data from the earnings conference calls of S\&P 1500 companies from 2015 to 2018 and aligned the two modalities to construct a dataset for financial risk prediction. \citet{yang2020html} developed a transformer-based multi-task architecture to learn the textual and audio features for predicting stock return volatility. \citet{yang2022numhtml} proposed two new training tasks (i.e., numeral category classification and magnitude comparison) for better capturing the numerical information for volatility forecasting. It is worth noting that the aforementioned studies mainly focus on the modeling of textual and audio modalities but ignore the valuable information on dialogue structure and company relationships. To address these limitations, \citet{ye2020financial} proposed a multi-round question-and-answer attention network (MRQA) to model the dialogue structure (e.g., the sentence importance and association). \citet{sawhney2020voltage} are among the first to exploit the inter-company relationships using the GCN model for financial risk prediction \citep{sawhney2020voltage}. However, they construct companies that held earnings conference calls on different dates in an undirected graph, causing temporal information leakage in which the model can utilize the information that occurs after the target event to make predictions. Differing from existing models, we exploit more dialogue structures by encoding position, utterance, speaker role, and Q\&A segments and propose a new method for constructing temporal company networks with no temporal information leakage.

\section{Methodology} \label{sec:method}
In this section, we first formulate our volatility forecasting problem and then elaborate on our proposed model.

\subsection{Problem Formulation}
We follow \citet{kogan2009predicting} to measure the financial risk using stock return volatility. In most cases, the higher the volatility, the riskier the stock. Formally, we define the stock return volatility $y_{[t,t+\tau]}$ from the trading day $t$ to $t+\tau$ as the standard deviation of returns $\sqrt{\sum_{i=0}^{\tau}(r_{t+i}-\overline{r})^2/\tau}$, where $r_t$ is the adjusted return of the given stock on the trading day $t$, $\overline{r}$ is the average of the adjusted returns from $t$ to $t+\tau$, and $\tau$ is the size of time window for calculating volatility. The adjusted stock return is defined as $r_{t} = p_{t} / p_{t-1} - 1$, where $p_{t}$ is the adjusted closing price of the stock on the trading day $t$. According to the post-announcement drift phenomenon, a stock's cumulative abnormal returns tend to drift in the direction of an earnings surprise for several weeks following an earnings announcement \citep{ball1968empirical}. Hence, we set the time windows $\tau$ of stock return volatility as 3, 7, and 15 days. We formulate our volatility forecasting problem as a supervised regression task. Specifically, given the sentence sequence $S_{i} = [s_{i, 1}, s_{i, 2}, \cdots, s_{i, N}]$ of the earnings conference call held by the company $c_i$ on the day $t$, we aim to predict the company's stock return volatility $y_{[t+1,t+\tau]}$ from the day $t+1$ to $t+\tau$.


\subsection{Proposed Model}
\begin{figure*}[htbp]
    \centering
    \includegraphics[width=17cm]{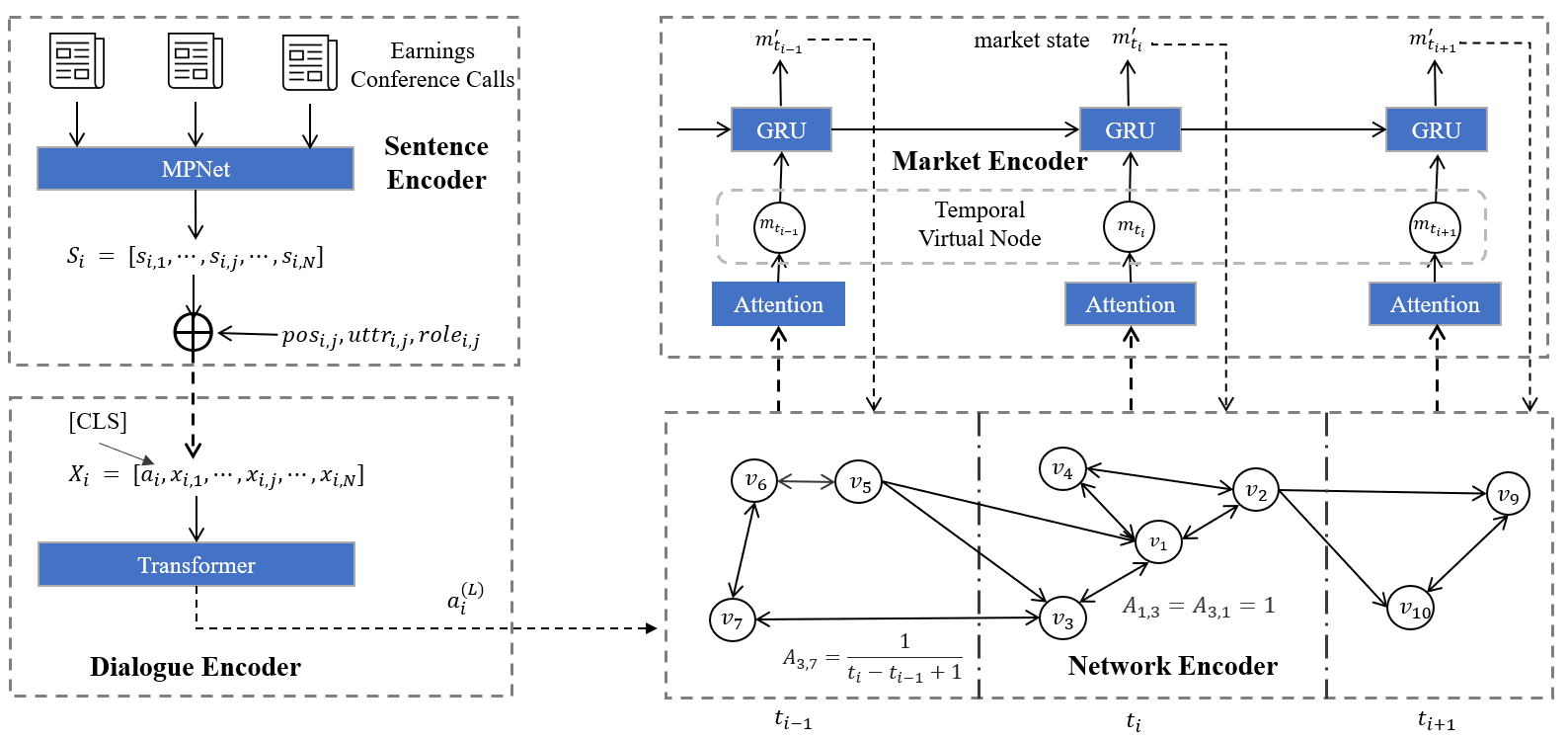}
    \caption{The Architecture of TVGNN}
    \label{fig:model}
\end{figure*}

As shown in Figure \ref{fig:model}, Our proposed TVGNN (Temporal Virtual Graph Neural Network) model contains three main modules, including:
\begin{enumerate}
\item Sentence Encoder: This module extracts the textual information of the earnings conference call and incorporates dialogue structural features to output a vector representation for each sentence.
\item Dialogue Encoder: This module captures the contextual relationship of sentences in earnings conference calls and encodes the entire dialogue into a vector.
\item Company Network Encoder: This is the key module of our model, consisting of company network construction, market encoder, and network encoder. First, we construct company networks based on company relationships, guaranteeing no temporal information leakage. The market encoder then models the market state at each time step. Finally, the network encoder employs a graph neural network to fuse all information and update the company representations.
\end{enumerate}
At last, the obtained company representations are fed into an output layer for the downstream task.

\subsubsection{Sentence Encoder}
We use a pre-trained model MPNet \citep{song2020mpnet} to encode sentences. The sequence of encoded sentences is denoted as $S_{i} = [\mathbf{s}_{i, 1}, \cdots, \mathbf{s}_{i, j}, \cdots, \mathbf{s}_{i, N}]$, where the $\mathbf {s}_{i, j}$ is the encoded vector of the $j$-th sentence in the earnings conference call of company $c_{i}$.
According to the position encoding in BERT \citep{Devlin2019BERTPO}, we use four structural embeddings to represent the dialogue structure of an  earnings conference call:
\begin{enumerate}
    \item Position embedding $pos_{i, j}$ encodes the order of sentences in the dialogue.
    \item Utterance embedding $uttr_{i, j}$ encodes to which utterance the sentences subordinate. All sentences spoken by a speaker at once are called an utterance.
    \item Role embedding $role_{i, j}$ encodes the speaker's role information of a given sentence. There are two roles in an earnings conference call: executives and analysts.
    \item Part embedding $part_{i, j}$ encodes in which parts the sentence appears. An earnings conference call consists of two parts: presentation and Q\&A.
\end{enumerate}

Finally, we concatenate the structural embeddings to the sentence vector $\mathbf{s}_{i, j}$:

\begin{equation}
    \mathbf{x}_{i, j} = \mathbf{s}_{i, j} \oplus pos_{i,j} \oplus uttr_{i, j} \oplus role_{i,j} \oplus part_{i, j}
\end{equation}

In this way, we obtain a sequence of sentence vectors $X_{i} = [\mathbf{x}_{i, 1}, \mathbf{x}_{i, 2}, \cdots, \mathbf{x}_{i, N}]$ that incorporates dialogue structural information, where each vector $\mathbf{x}_{i, j} \in \mathbb {R}^{d}$.

\subsubsection{Dialogue Encoder}
The key to text modeling is to model the contextual relationships. To obtain contextual information, we update sentence vectors with a Transformer encoder \citep{vaswani2017attention}. Following BERT, we add a trainable [CLS] vector $\mathbf{a}$ to the input sequence $X_{i}$ to obtain the representation of a given earnings conference call. Then the input sequence becomes $X_{i} = [\mathbf{a}, \mathbf{x}_{i, 1}, \mathbf{x}_{i, 2}, \cdots, \mathbf{x}_{i, N}]$. After $L$-layer Transformer encoders, we take $\mathbf{a}^{(L)}$ in the output sequence as the representation of the earnings conference call, denoted as $\mathbf{v}_{i}$.

\subsubsection{Company Network Encoder}
The company network encoder contains two submodules: a market encoder and a network encoder that model market state and company networks, respectively.

In this section, we first describe how to construct company networks. Then, we introduce how to apply the market encoder and the network encoder on the constructed company network to model the market state and the company relationships.

\textbf{Company Network Construction}
Assume there are $M$ companies holding their earnings conference calls in a given quarter (natural quarter), we treat each company as a node in the graph, and the node $v_{i}$ corresponds to company $c_{i}$. If there are some kinds of relationships existing between $c_{i}$ and $c_{j}$ and the date $t_{c_{j}}$ when $c_{i}$ holds its earnings conference call in the given quarter is not earlier than $t_{c_{j}}$, a weighted directed edge is connected between nodes $v_{i}$ and $v_{j}$. The weight of the edge is defined as follows.

\begin{equation}
    \small
    \mathbf{A}_{i,j}=
    \begin{cases}
    \frac{1}{(t_{c_i} - t_{c_j} + 1)} & \text{if $v_{i}$ is connected with $v_{j}$ and $t_{c_i}-t_{c_j} >= 0$} \\
    0 & \text{otherwise}
    \end{cases}
    \label{eq:a}
\end{equation}

The weight defined in Equation (\ref{eq:a}) ensures that the weight of the edge between two nodes decreases as the interval between the two corresponding companies' earnings conference calls increases. 


The constructed company network is a directed static graph. The network's directed edges prevent temporal information leakage, i.e., information can only flow from temporally preceding nodes to temporally following nodes, which is critical for a temporal prediction task.

Finally, we assign $\mathbf{c}_{i}$ obtained from the dialogue encoder as the initial embedding of the node $v_{i}$.

\textbf{Market Encoder}
The capital asset pricing model points out that the return of an asset consists of the risk-free return of the market and the return of the asset \citep{blume1973new}. Therefore, when predicting the risk of a given company, we should consider the impact of both the events (earnings conference calls) of the company and the market state. Thus, we design the market encoder to model the market state $\mathbf{m}_{t_{i}}$ at day $t_{i}$. 


Suppose there are $T$ different dates $[t_{1}, \cdots, t_{i}, \cdots, t_{T}]$ for holding an earnings conference call in a given quarter. The dates are sorted in chronological order. The set of companies holding earnings conference call at $t_{i}$ are denoted as $\mathcal{V}_{t_{i}}$.

The market state of $t_{i}$ can be represented as the sum of all events (earnings conference calls) happening in the market at $t_{i}$. Thus, we can calculate the market state $\mathbf{m}_{t_{i}}$ by a global attention module:

\begin{align}
    \mathbf{m}_{t_{i}} &= \sum_{v_{j} \in V_{t_{i}}} \beta_{j} \mathbf{v}_{j} \\
    \mathbf{k}_{j} &= \mathbf{v}_{j} \mathbf{w}^{T}_{k} \\
    e_{j} &= \frac{ \mathbf{w}_{q} \mathbf{k}^{T}_{j} }{\sqrt{d}} \\
    \beta_{j} &= \frac{e_{j}}{\sum_{v_{u} \in \mathcal{V}_{t_{i}}} e_{u}}
\end{align}

where $\beta_{j}$ is the attention sore of node $v_{j} \in \mathcal{V}_{t_{i}}$, $\mathbf{w}_{q}$ is a trainable parameter.


The market state $\mathbf{m}_{t_{i}}$ is affected not only by the events at day $t_{i}$, but also by the market state of the past. Thus, we use a Gated Recurrent Unit Network (GRU) \citep{li2015gated} to model the historial market state. Since the larger the time interval, the weaker the effect of the previous market state on the current market state, we define a coefficient $\Delta_{t_{i}}=\sigma( \frac{w_{d}}{t_{i} - t_{i-1} + 1} )$ which represents the time interval between $t_{i}$ and $t_{i-1}$, to adjust the value of the reset gate in the GRU:

\begin{align}
\mathbf{z}_{t_{i}} &= \sigma(\mathbf{m}_{t_{i}} \mathbf{w}_{1}^{T} + \mathbf{a}_{t_{i-1}} \mathbf{w}_{2}^{T} ) \label{eq:gru1} \\
\mathbf{r}_{t_{i}} &= \sigma(\mathbf{m}_{t_{i}} \mathbf{w}_{1}^{T} + \mathbf{a}_{t_{i-1}} \mathbf{w}_{2}^{T} ) \label{eq:gru2} \\
\tilde{\mathbf{a}}_{t_{i}} &= \tanh (\mathbf{m}_{t_{i}} \mathbf{w}_{1}^{T} + (\Delta_{t_{i}} \circ \mathbf{r}_{t_{i}} \circ \mathbf{h}_{t_{i- 1}}) \mathbf{w}_{2}^{T}) \label{eq:gru4} \\
\mathbf{a}_{t_{i}} &= (1-\mathbf{z}_{t_{i}}) \circ \mathbf{a}_{t_{i-1}}+\mathbf{z}_{t_{i}} \circ \tilde{\mathbf{a}}_{t_{i}} \label{eq:gru5 } \\
\mathbf{m}^{\prime}_{t_{i}} &= \mathbf{a}_{t_{i}} \mathbf{w}_{a}^{T} + b_{a}
\end{align}
where $\sigma$ is an activation function.

\textbf{Network Encoder}
We use a graph attention network with symmetric normalized adjacency matrix \citep{wang2021bag} to update node embedding so that each company node can capture the topology of the company network. Assume using $L$-layer graph attention networks, the update formula of the company node embedding at the $l$-layer is as follows.

\begin{align}
\mathbf{g}^{(l-1)}_{i} &= \mathbf{v}^{(l-1)}_{i} + \mathbf{m}^{(l) \prime}_{t(c_{i})} \label{eq:gat1} \\
\mathbf{v}^{(l)}_{i} &= \sigma(\sum_{v_{j} \in \mathcal{N}(v_i)} \frac{\gamma_{ij}}{\tilde{D}_{ij}} \mathbf{g}^{(l-1)}_{j} \mathbf{w}^{T}_ {0}+\mathbf{g}^{(l-1)}_{i} \mathbf{w}_{1}^{T}) \label{eq:gat2}
\end{align}

where equation (\ref{eq:gat1}) adds the company node embedding  $\mathbf{v}^{(l-1)}_{i}$ and corresponding market state $\mathbf{m^{(l)}_{t(c_{i})}}$ to obtain the input vector $\mathbf {g}^{(l-1)}_{i}$ for the $l$-th layer of graph attention networks. $\gamma_{ij}$ is the attention score between $v_{i}$ and $v_{j}$, calculated by

\begin{equation}
\gamma_{i j}=\frac{\exp \{ \operatorname{LeakyReLU}\{[(\mathbf{v}_{i} \oplus \mathbf{v}_{j}) \mathbf{w}_{1}^{T}] (\mathbf{e}_{i j} \mathbf{w}_{2}^{T}) \}\} }{\sum_{r \in \mathcal{N}(v_{i})} \exp \{  \operatorname{LeakyReLU} \{ [(\mathbf{v}_{i} \oplus \mathbf{v}_{r}) \mathbf{w}_{1}^{T}] (\mathbf{e}_{i r} \mathbf{w}_{2}^{T}) \} \} }
\end{equation}

where $\mathbf{w}_{1}^{T}$ and $\mathbf{w}_{2}^{T}$ are trainable parameters. Compared to the vanilla graph attention network, we introduce the edge feature $\mathbf{e}_{ij}$ into the formula of attention score. Because the edge features commonly measure the strength of connections, we map it to a scalar coefficient to adjust the attention scores.

Finally, the company network encoder combines the market encoder and network encoder. The constructed company network and the company embedding are input into the company network encoder. For the $i$th-layer company network encoder, it alternately uses a market encoder and a network encoder to update the company node embedding. In fact, the company encoder can be seen as a model that adds temporal virtual nodes to the company nodes and then uses extended GRU and GAT to update node embedding.

\subsubsection{Output Layer}
We use a multilayer perceptron as the output layer to predict stock return volatility:
\begin{equation}
    \hat{y}_{i} = \sigma(\mathbf{v}^{(L)}_{i} \mathbf{w}^T_1 + b_1)\mathbf{w}_2 + b_2
\end{equation}
where $\mathbf{w}_1, \mathbf{w}_2, b_1, b_2$ are trainable parameters, $\sigma$ is an activation function.

\section{Experiments}\label{sec:experiments}
In this section, we first describe our dataset and baselines and then present the experimental results.

\subsection{Dataset}
To evaluate model performance, we construct a new conference call dialogue dataset with additional information on dialogue structures (e.g., position, untterance, speaker role and segment information) and company networks. Specifically, we collect the earnings conference call transcripts of the S\&P500 companies from 2008 to 2019 from the website of SeekingAlpha, and use the Text-based Network Industry Classifications (TNIC) dataset \citep{hoberg2016text} for constructing the compnay network. TNIC computes company pairwise similarity scores based on the product descriptions in 10-K files \citep{hoberg2016text}. We add edges for two companies with a similarity score greater than 0.15. Since the TNIC dataset is updated annually based on the latest 10-K files, we use the TNIC dataset from the previous year to extract the company relationships to avoid temporal information leakage.



We split the dataset by time, using samples before 2016 as the training set, samples from 2016 as the validation set, and samples after 2016 as the test set. For a natural quarter, all samples are constructed as a company network, , with each node labeled with the corresponding company's stock return volatility.

\begin{table*}[htp!]
\centering
\caption{Experimental Results}
\label{tab:main_results}
\begin{tabular}{@{}cccccccc@{}}
\toprule
Model & $\overline{\text{MSE}}$ & $\text{MSE}_3$  & $\text{MSE}_7$  & $\text{MSE}_{15}$ & $R^2_3$ & $R^2_7$ & $R^2_{15}$ \\ \midrule
$v_{past}$ & 0.5843 & 1.1336 & 0.4026 & 0.2167 &        &         &         \\
HAN        & 0.6698 & 0.8911 & 0.6199 & 0.4983 & 0.2139 & -0.5397 & -1.2995 \\
MDRM       & 0.6135 & 0.7958 & 0.5560 & 0.4888 & 0.2980 & -0.3810 & -1.2557 \\
ProFET     & 0.5098 & 0.7527 & 0.4378 & 0.3390 & 0.3360 & -0.0874 & -0.5644 \\
MRQA       & 0.4637 & 0.7141 & 0.3796 & 0.2974 & 0.3701 & 0.0571  & -0.3724 \\
HTML       & 0.4156 & 0.6969 & 0.3097 & 0.2401 & 0.3852 & 0.2308  & -0.1080 \\
TVGNN & \textbf{0.3605}         & \textbf{0.5980} & \textbf{0.2818} & \textbf{0.2017}   & 0.4725  & 0.3000  & 0.0692     \\ \bottomrule
\end{tabular}
\end{table*}

\subsection{Baselines}
In order to verify the effectiveness of TVGNN, we compare it with baselines proposed in recent related studies. The baselines are shown below.

\begin{enumerate}
    \item $v_{past}$ \citep{kogan2009predicting}. $v_{past}$ is a simple but very effective benchmark model, which directly uses the return volatility $v_{[t-\tau, t-1]}$ as the  prediction $v_{[t+1, t+\tau]}$, without using any other information.
    \item HAN \citep{yang2016hierarchical}. The model is a widely used long document encoder that employs two BiGRU to capture contextual relationships at the word and sentence levels, as well as a simple attention module to obtain encoding at each level. The obtained document encoding is then used for downstream tasks.
    \item ProFET \citep{theil2019profet}. This model simply considers the dialogue structure of earnings conference calls and  respectively models the presentation and Q\&A part by a BiLSTM and an attention module. We use the text modeling part of ProFET for experiments.
    \item MDRM \citep{qin2019what}. The model employs a BiLSTM to extract textual and audio features from earnings conference calls, which are then fused by another BiLSTM to predict the company's risk. We use the text modeling part of this model for experiments.
    \item HTML \citep{yang2020html}. The model repectively uses a pre-trained language model to extract word encoding and Praat to extract audio features. The features are then fused to obtain a multi-modal encoding at the sentence level. The model is trained in a multi-task framework with an auxiliary task of predicting the return of the company's stock on a given day. We use the text modeling  and the multi-task part of the model for experiments.
    \item MRQA \citep{ye2020financial}. The model uses a BiLSTM to encode the textual features of earnings conference calls. It uses a reinforced sentence selector to select important sentences in the Q\&A, and a reinforced bidirectional attention network to capture the interaction between questions and answers. The model directly models the dialogue structure of earnings conference calls.
\end{enumerate}

\subsection{Experimental Settings}

We use the Mean Square Error (MSE) as the loss function and evaluation metric:
\begin{equation}
    \text{MSE} = \sum_{i}^{M}(\hat{y}_{i} - y_{i})^2 / M
\end{equation}
where $M$ is the number of samples. For robustness, we also use the $R^2$ metric to measure the performance improvement over the simplest baseline $v_{past}$:
\begin{equation}
    R^{2}=1-\text{MSE} / \text{MSE}_{v_{past}}
\end{equation}

We use the Adam optimizer for training models \citep{kingma2014adam}, and tune the hyperparameters of our TVGNN model on the validation set using the MSE metric. The tuned hyperparameters are as follows: the learning rate is 5e-4, the weight decay is 1e-7, the hidden state is 64, the number of layers of dialogue encoder is 2, the number of attention heads of dialogue encoder is 8, the number of layers of company network is 3, and the number of attention head of company network encoder is 1. We use the default hyperparameters for all baselines as in their original papers and stop the training if the MSE score on the validation set does not decrease in 10 epochs.

\subsection{Experimental Results}
Table \ref{tab:main_results} shows the result of the comparison experiment, we can see that TVGNN performs best in the company risk prediction tasks for the three time windows. Compared to the best baseline models, TVGNN respectively improves 14.19\%, 9.01\%, and 6.92\% for $\tau=3, 7, 15$. On the overall performance $\overline{\text{MSE}}$, TVGNN improves 13.26\% at the basis of the best baseline HTML, demonstrating the effectiveness of TVGNN.
Additionally, as the time window gets longer, the difference between $v past$ and other baselines gets smaller. For $\tau=15$, all baselines are worse than $v_{past}$, indicating that the additional effect of the earnings conference calls transcripts gradually diminishes as the time window becomes longer, which is consistent with the Earning Momentum \citep{ball1968empirical}. The impact of a company's earnings information on its stock price fades over time. In our experiment, the effect lasts no longer than 15 days.

\subsection{Supplementary Analysis}\label{sec:supplementary}
We further conduct supplementary analysis to examine our model's key modules, nowcasting, and model interpretability.

\subsubsection{Ablation Study}
We conduct the ablation study to examine the effectiveness of our TVGNN model's four modules, including the sentence encoder, dialogue encoder, market encoder, and network encoder. The results of the ablation study are shown in Table 1, in which the "+" symbol indicates that the model variant contains the corresponding module. Our main findings are:

\begin{enumerate}
    \item Adding each module can improve the model's performance in company risk prediction, demonstrating the effectiveness of the four modules in TVGNN.
    \item It is difficult for a model to improve on all three tasks without introducing new data. When $tau=3$ and $tau=7$, variant 2 outperforms variant 1, but when $tau=15$, it outperforms variant 1. While variant 3 outperforms variant 2 on all three tasks by incorporating company networks.
    \item Modeling the market state can improve the model's performance on the task $\tau=15$. TVGNN with the market encoder outperforms variant 3 on $tau=15$.
\end{enumerate}

\begin{table*}[htp!]
\centering
\caption{Ablation Experimental Result}
\label{tab:ablation}
\begin{tabular}{@{}ccccccccc@{}}
\toprule
Variant &
  \begin{tabular}[c]{@{}c@{}}Sentence\\ Encoder\end{tabular} &
  \begin{tabular}[c]{@{}c@{}}Dialogue   \\ Encoder\end{tabular} &
  \begin{tabular}[c]{@{}c@{}}Market   \\ Encoder\end{tabular} &
  \begin{tabular}[c]{@{}c@{}}Network   \\ Encoder\end{tabular} &
  $\overline{\text{MSE}}$ &
  $\text{MSE}_3$ &
  $\text{MSE}_7$ &
  $\text{MSE}_{15}$ \\ \midrule
1     & + &   &   &   & 0.4162 & 0.6616 & 0.3354 & 0.2517 \\
2     & + & + &   &   & 0.4116 & 0.6451 & 0.3193 & 0.2704 \\
3     & + & + & + &   & 0.3700 & 0.6065 & 0.2773 & 0.2261 \\
TVGNN & + & + & + & + & 0.3605 & 0.5980 & 0.2818 & 0.2017 \\ \bottomrule
\end{tabular}
\end{table*}

\subsubsection{Nowcasting}
In the previous analysis, we constructed the static company graph and updated our TVGNN model on a quarterly basis. But in the real-world application, we may need to perform the prediction immediately. Hence, to better use our model in practice, we propose two learning approaches: TVGNN-T and TVGNN-T (fine-tune). For the experiment, we first split the sample in the first quarter of 2017 into training/validation/testing sets in chronological order with the ratio of 7:1:2 and trained the following models to forecast volatility on the split testing set.

\begin{enumerate}
\item TVGNN: This model is is the same as the one in Table \ref{tab:main_results}, using the training sets before 2017 to train.
\item TVGNN-T: This model is trained and validated using the splitted training and validation sets in the first quarter of 2017.
\item TVGNN-T (fine-tune): Based on TVGNN in 1, this model is trained using the splitted training sets in the first quarter of 2017.
\end{enumerate}
As can be seen in Table \ref{tab:trans}, the TVGNN-T model performs worse because it is trained on a small sample, while the TVGNN and TVGNN-T (fine-tune) models benefit from the extensive training sample. Also, the TVGNN-T (fine-tune) outperforms TVGNN for nowcasting because it uses the smaller but more timely samples during the first quarter of 2017.

\begin{table}[htp]
\centering
\caption{Nowcasting}
\label{tab:trans}
\begin{tabular}{@{}ccccc@{}}
\toprule
Model               & $\overline{\text{MSE}}$ & $\text{MSE}_3$ & $\text{MSE}_7$ & $\text{MSE}_{15}$ \\ \midrule
TVGNN               & 0.3819                  & 0.5694         & 0.3103         & 0.2659            \\
TVGNN-T             & 0.5847                  & 0.7721         & 0.5962         & 0.3857            \\
TVGNN-T (fine-tune) & 0.3708                  & 0.5586         & 0.2961         & 0.2576            \\ \bottomrule
\end{tabular}
\end{table}

\subsubsection{Interpretability}
To examine the model interpretability, we use the GNNExplainer to identify the important subgraph that influences the prediction \citep{ying2019gnnexplainer}. We conducted a case study of Oracle, whose cloud business soared and increased its dividend in the first quarter of 2017.

\begin{table*}[htp]
\centering
\caption{Selected Companies for Case Study}
\label{tab:company}
\begin{tabular}{@{}cccc@{}}
\toprule
Symbol & Security.        & GICS   Sub-Industry      & $v_{\tau=3}$ \\ \midrule
ORCL   & Oracle           & Application Software     & -3.2365      \\
CTXS   & Citrix           & Application Software     & -3.1539      \\
FFIV   & F5               & Communications Equipment & -2.9807      \\
FTNT   & Fortinet         & Systems Software         & -2.5418      \\
JNPR   & Juniper Networks & Communications Equipment & -3.6766      \\
CSCO   & Cisco            & Communications Equipment & -4.6155      \\
HPE & Hewlett Packard Enterprise & Technology Hardware, Storage \&   Peripherals & -3.0791 \\
ANSS   & Ansys            & Application Software     & -4.1327      \\
PTC    & PTC              & Application Software     & -3.7511      \\
MSFT   & Microsoft        & Systems Software         & -5.3507      \\
CRM    & Salesforce       & Application Software     & -3.7586      \\\bottomrule
\end{tabular}
\end{table*}

\begin{figure}[htp]
    \centering
    \includegraphics[width=9.5cm]{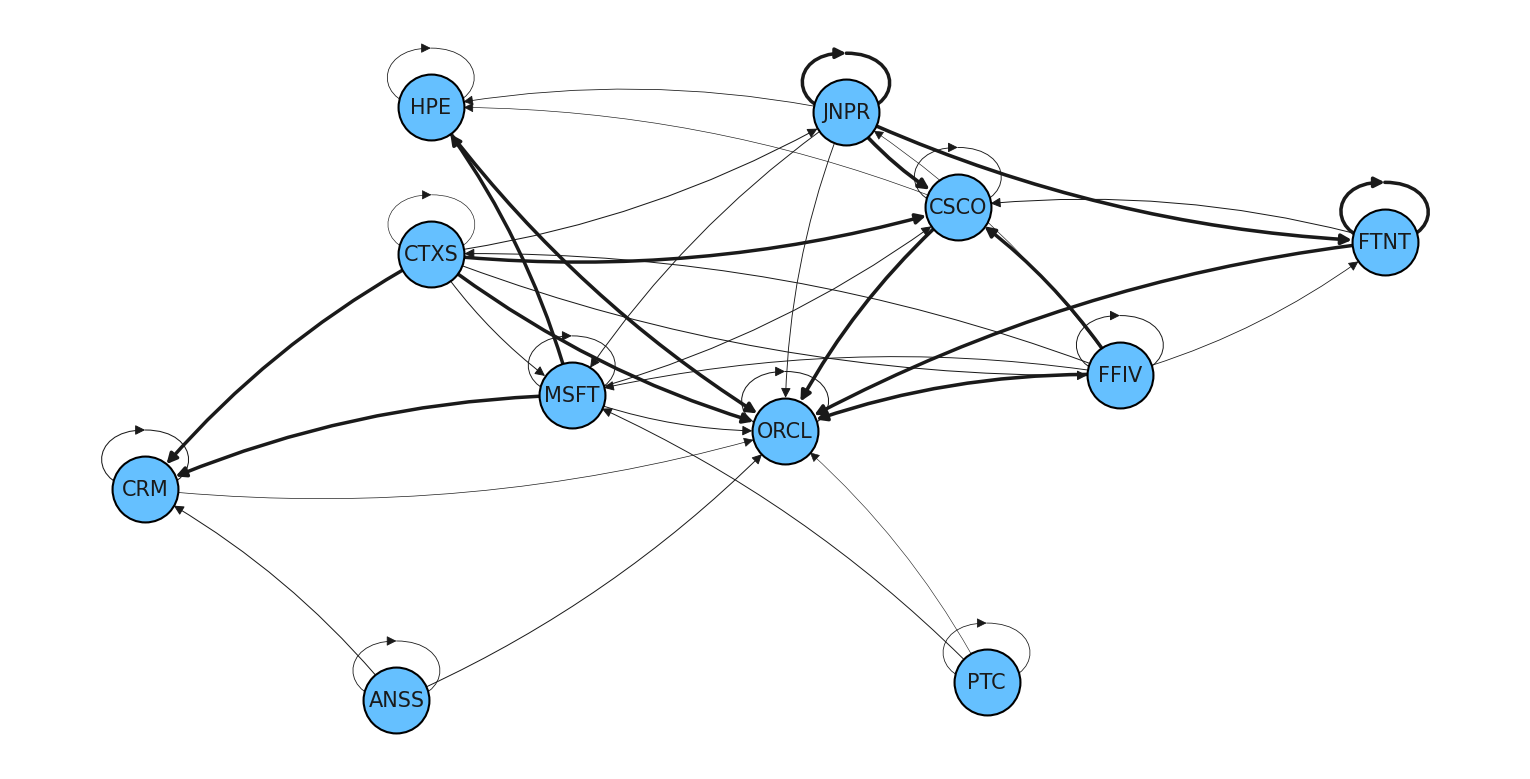}
    \caption{Case Study}
    \label{fig:explainer}
\end{figure}

The important 1-hop subgraph of ORCL (Oracle) identified by GNNExplainer is shown in Figure \ref{fig:explainer}. The details of the nodes are shown in Table \ref{tab:company}. The edge width in the subgraph represents the edge's important score. Despite the fact that all of the companies in the subgraph are in the information technology sector, not all of them have a significant impact on the risk of other companies. The prediction of our TVGNN $\hat{y}_{\tau=3}$ for ORCL (Cisco) is mainly influenced by CTXS, HPE, CSCO, FTNT, and FFIV. All five companies are Oracle partners, and they all sell Oracle products in their cloud services. Furthermore, each node in the graph has a self-loop, which represents the impact of the company's event, i.e., the earnings conference call, on the risk of the company. We can see that the risk of JNPR (Juniper Networks) and FTNT (Fortinet) is influenced more by their earnings conference calls, whereas the rest is influenced more by other companies in the company network. This is because JNPR and FTNT released some shocking financial information during their earnings conference calls. JNPR posted a disappointing profit outlook, causing its stock to plunge 7.5\% after the earnings conference call. FTNT's stock, on the other hand, jumped 10.8\% after its earnings conference call, as its performance in 2016 gave investors confidence in its 2017 performance.


\section{Conclusion}\label{sec:conclusion}
In this paper, we propose a model called TVGNN for volatility forecasting by jointly modeling conference call dialogues and company networks. We make methodological contributions by designing a new method to construct company networks and developing a new model based on graph neural networks to model earnings conference calls, market state, and company networks. Empirical results on our constructed dataset demonstrate the superiority of our proposed model over competitive baselines from the extant literature. We also conduct supplementary analysis to examine our model's key modules, nowcasting, and model interpretability.


\bibliography{paper}

\appendix

\section{Dataset Statistics}\label{sec:dataset}

\begin{figure}[htbp]
\centering
\subfigure[$\tau=3$]{
    \centering 
    \includegraphics[width=5cm]{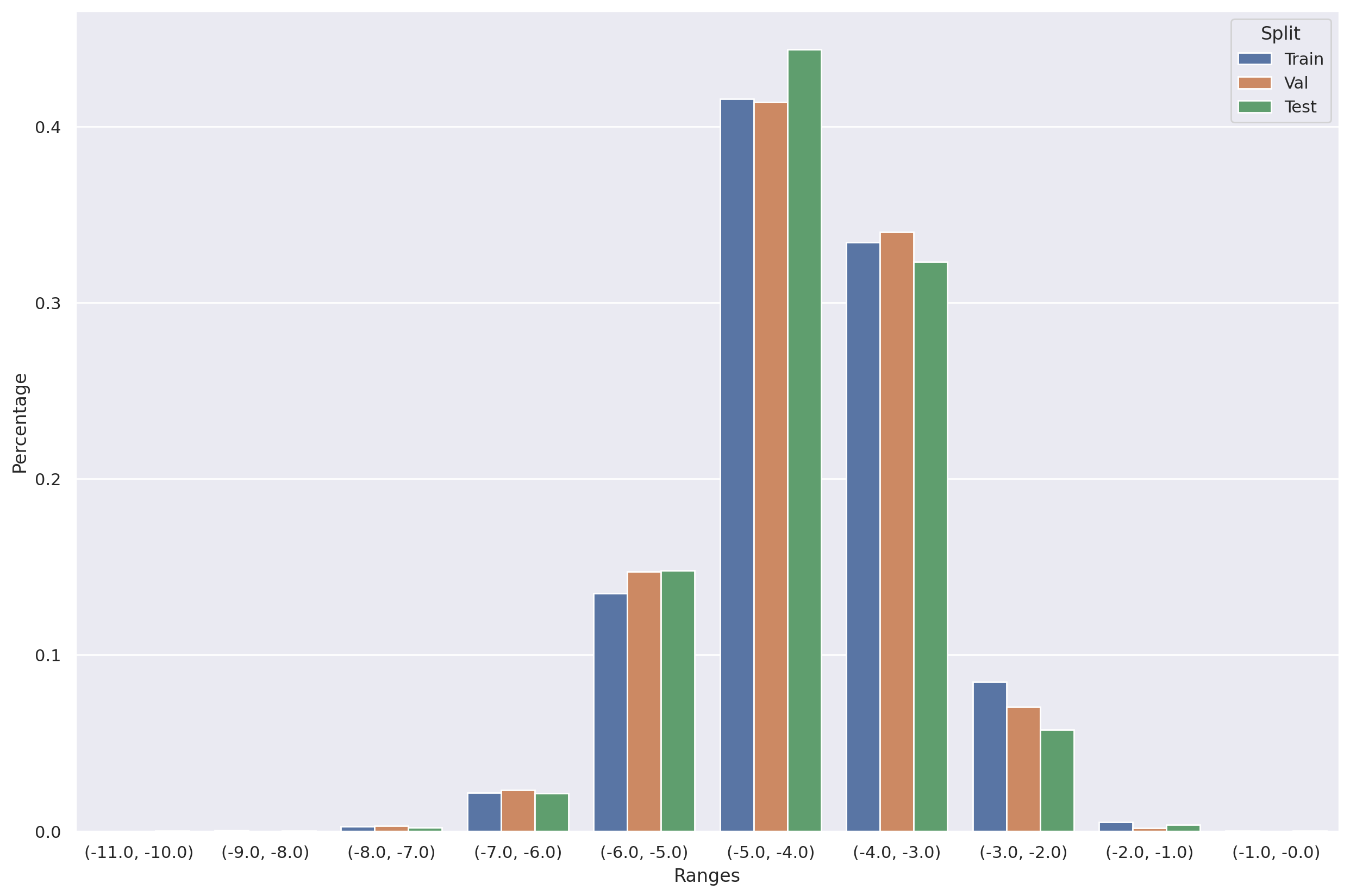}
}
\subfigure[$\tau=7$]{
    \centering
    \includegraphics[width=5cm]{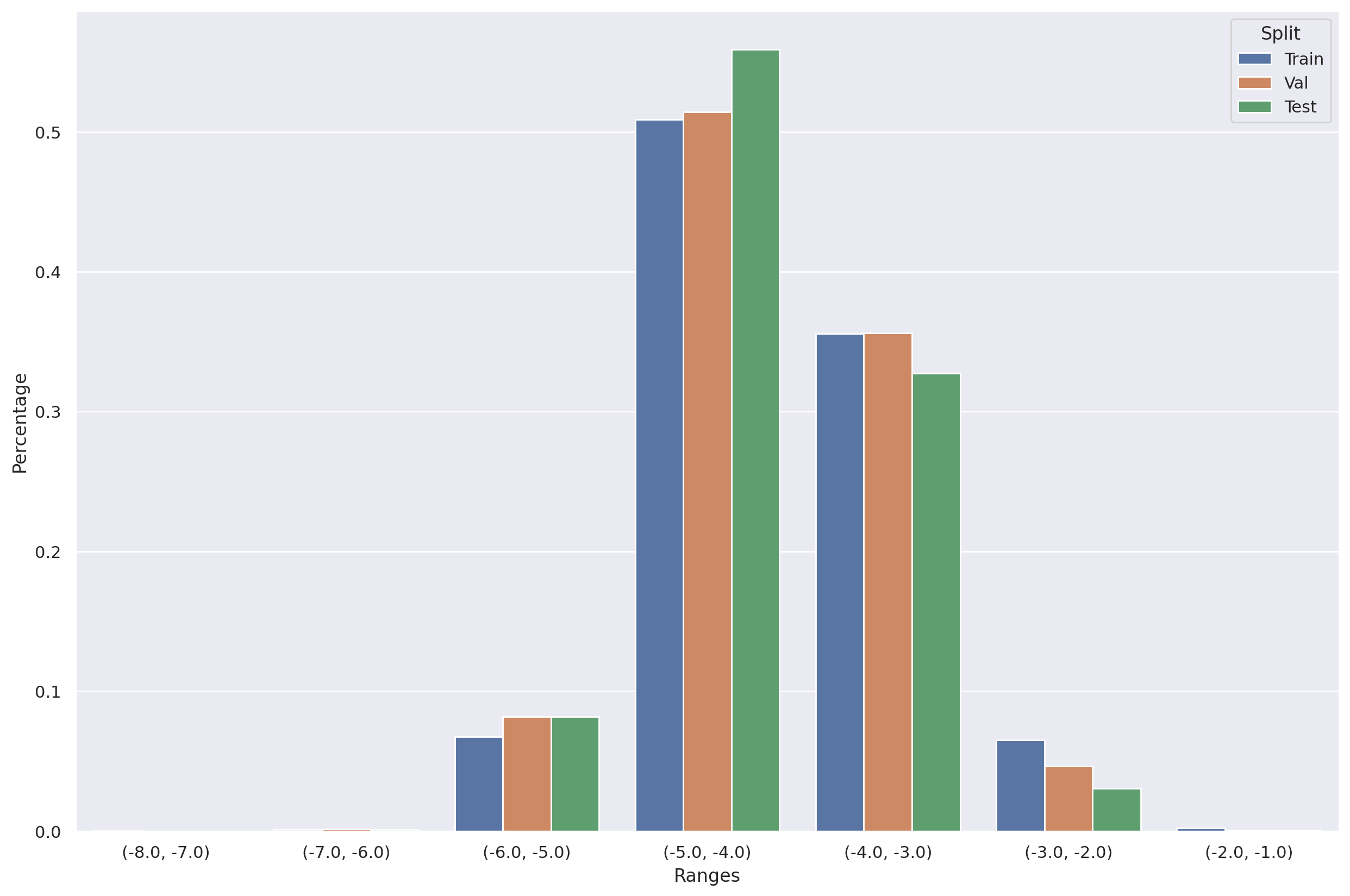}
}
\subfigure[$\tau=15$]{
    \centering
    \includegraphics[width=5cm]{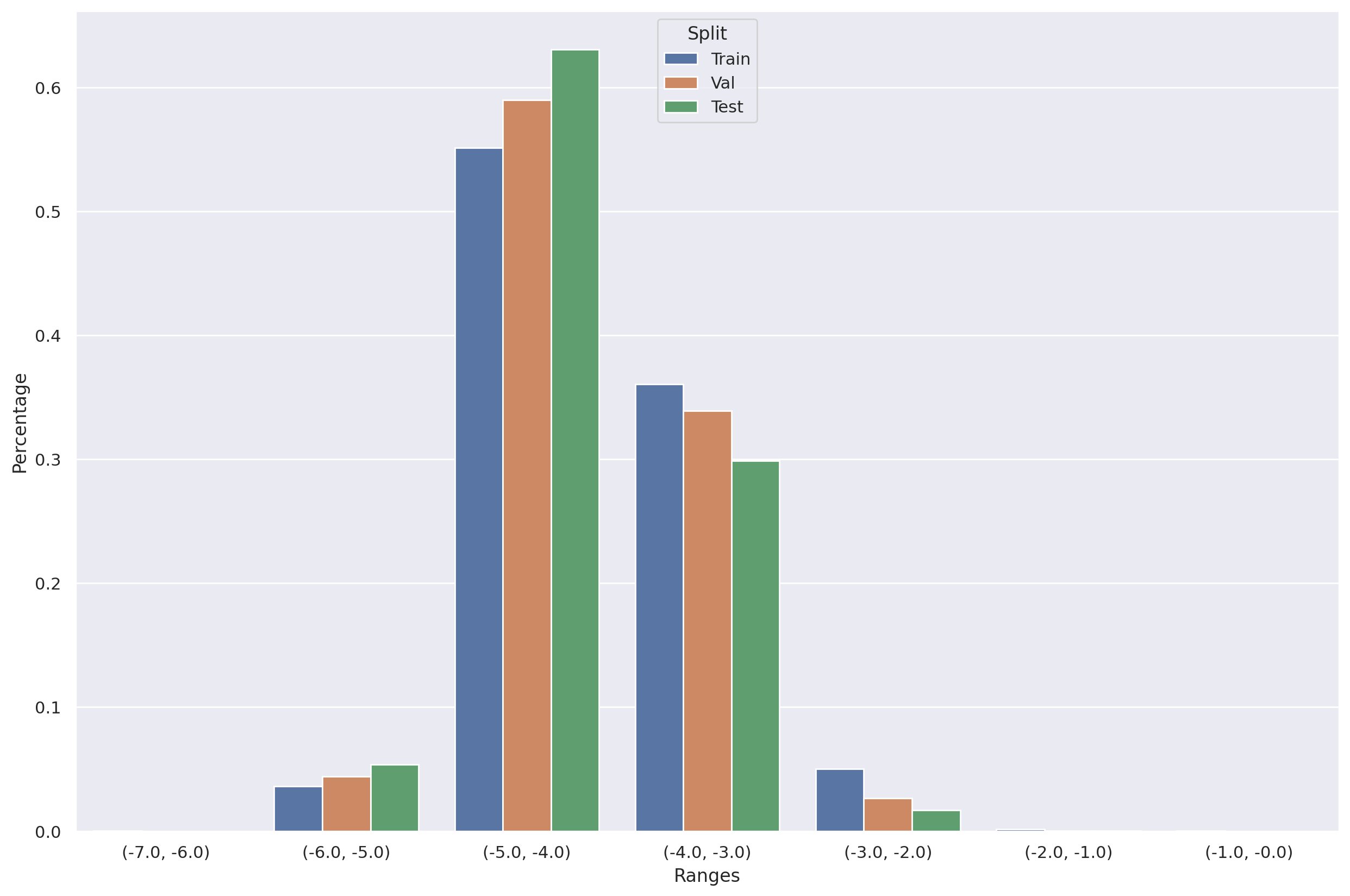}
}
\caption{The Distribution of Labels. The distribution of labels in different splits is consistent.}
\label{fig:label}
\end{figure}

Table \ref{tab:dataset} shows the descriptive statistics of our dataset. Figure \ref{fig:label} shows the volatility distribution in the training, validation, and testing sets.

\begin{longtable}[c]{@{}ccccc@{}}
\caption{Dataset Statistics}
\label{tab:dataset}\\
\toprule
Quarterly & Average \#Utterance & Average \#Sentences & \#Nodes (Companies) & \#Edges   \\* \midrule
\endfirsthead

\toprule
Quarterly & Average \#Utterance & Average \#Sentences & \#Nodes (Companies) & \#Edge   \\* \midrule
\endhead
\bottomrule
\endfoot
\endlastfoot
2008Q1 & 95.90 & 162.63 & 559     & 4159 \\
2008Q2 & 87.87 & 148.37 & 566     & 4535 \\
2008Q3 & 88.81 & 156.59 & 502     & 3239 \\
2008Q4 & 87.08 & 166.66 & 553     & 4238 \\
2009Q1 & 86.56 & 165.87 & 570     & 4768 \\
2009Q2 & 81.68 & 159.23 & 514     & 4167 \\
2009Q3 & 78.69 & 155.77 & 515     & 4124 \\
2009Q4 & 79.07 & 153.47 & 481     & 3940 \\
2010Q1 & 78.19 & 149.72 & 511     & 3965 \\
2010Q2 & 78.09 & 139.39 & 466     & 3613 \\
2010Q3 & 79.00 & 138.47 & 378     & 2453 \\
2010Q4 & 76.48 & 135.69 & 362     & 2149 \\
2011Q1 & 74.39 & 126.01 & 479     & 3922 \\
2011Q2 & 76.27 & 121.81 & 478     & 3960 \\
2011Q3 & 73.53 & 119.81 & 575     & 5131 \\
2011Q4 & 73.34 & 120.33 & 572     & 5166 \\
2012Q1 & 73.34 & 127.89 & 578     & 5160 \\
2012Q2 & 69.17 & 120.75 & 577     & 5058 \\
2012Q3 & 72.56 & 123.17 & 596     & 5459 \\
2012Q4 & 69.11 & 125.97 & 579     & 4980 \\
2013Q1 & 58.89 & 108.21 & 607     & 5534 \\
2013Q2 & 61.78 & 112.30 & 631     & 5643 \\
2013Q3 & 60.71 & 110.79 & 627     & 5760 \\
2013Q4 & 63.40 & 115.59 & 622     & 5634 \\
2014Q1 & 68.05 & 125.27 & 624     & 5610 \\
2014Q2 & 69.88 & 123.60 & 615     & 5581 \\
2014Q3 & 69.78 & 125.75 & 617     & 5304 \\
2014Q4 & 68.88 & 126.05 & 608     & 5205 \\
2015Q1 & 73.38 & 138.32 & 614     & 5379 \\
2015Q2 & 74.42 & 137.20 & 593     & 5170 \\
2015Q3 & 78.00 & 138.48 & 615     & 5600 \\
2015Q4 & 77.58 & 140.49 & 593     & 4996 \\
2016Q1 & 78.03 & 147.08 & 590     & 5162 \\
2016Q2 & 77.02 & 137.34 & 589     & 5330 \\
2016Q3 & 77.22 & 139.42 & 566     & 4825 \\
2016Q4 & 76.18 & 139.45 & 581     & 5115 \\
2017Q1 & 77.79 & 144.41 & 594     & 4669 \\
2017Q2 & 75.00 & 135.30 & 578     & 4587 \\
2017Q3 & 71.62 & 133.09 & 569     & 4597 \\
2017Q4 & 73.36 & 139.12 & 561     & 4457 \\
2018Q1 & 68.28 & 139.41 & 587     & 4583 \\
2018Q2 & 71.29 & 134.49 & 588     & 4695 \\
2018Q3 & 69.55 & 131.77 & 572     & 4454 \\
2018Q4 & 70.60 & 134.98 & 573     & 4450 \\
2019Q1 & 67.51 & 147.38 & 555     & 4017 \\
2019Q2 & 66.14 & 141.53 & 537     & 4032 \\
2019Q3 & 66.16 & 145.75 & 497     & 3277 \\
2019Q4 & 64.15 & 143.60 & 493     & 3475 \\* \bottomrule
\end{longtable}

\end{document}